\documentclass{article}
\usepackage{spconf,extra_styles,amsmath,graphicx}
\usepackage{mathtools}
\usepackage{bbm}
\usepackage{makecell} 
\usepackage{threeparttable}
\usepackage{pifont}
\usepackage{multirow}
\usepackage{caption}
\usepackage{booktabs}
\usepackage{color}

\usepackage[pagebackref,breaklinks,colorlinks]{hyperref}
\title{TalkNCE: Improving Active Speaker Detection \\ 
with Talk-aware Contrastive Learning} % talking-aware? sync-aware? talk-aware?
\name{
    \begin{tabular}{c}
        Chaeyoung Jung$^{1*}$, Suyeon Lee$^{1*}$, Kihyun Nam$^{1}$, Kyeongha Rho$^{1}$, \\ You Jin Kim$^{2}$, Youngjoon Jang$^{1}$, Joon Son Chung$^{1}$
    \end{tabular}
    \thanks{$^*$ These authors contributed equally.}
}

\address{$^1$Korea Advanced Institute of Science and Technology, South Korea \\
$^2$Naver Cloud Corporation, South Korea}
\begin{document}
%\ninept
%\fontsize{9}{1}
%
\maketitle
\begin{abstract}
The goal of this work is Active Speaker Detection (ASD), a task to determine whether a person is speaking or not in a series of video frames. 
Previous works have dealt with the task by exploring network architectures while learning effective representations have been less explored. 
In this work, we propose TalkNCE, a novel talk-aware contrastive loss. The loss is only applied to part of the full segments where a person on the screen is actually speaking.
This encourages the model to learn effective representations through the natural correspondence of speech and facial movements.
Our loss can be jointly optimized with the existing objectives for training ASD models without the need for additional supervision or training data.
The experiments demonstrate that our loss can be easily integrated into the existing ASD frameworks, improving their performance.
Our method achieves state-of-the-art performances on AVA-ActiveSpeaker and ASW datasets. 

\end{abstract}

\section{Introduction}
\label{sec:intro}

% Figure 1
\begin{figure}[!t]
  \centering
  \includegraphics[width=7cm]{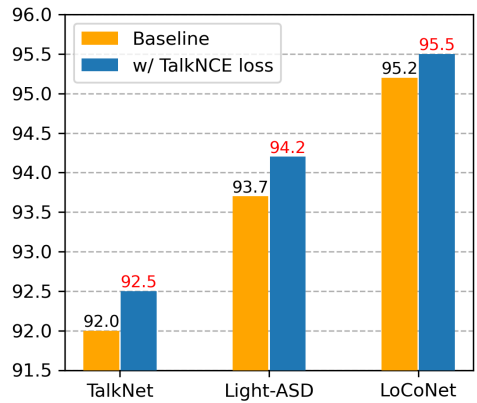}
\vspace{-3mm}
\caption{Comparison of performances on the validation set of AVA-ActiveSpeaker~\cite{roth2020ava}. 
Our TalkNCE loss improves the performance of existing ASD models~\cite{talknet,lightASD,loconet} consistently.}
% Using our TalkNCE loss, the performance of the existing ASD models \cite{talknet, lightASD, loconet} consistently increases.}
\vspace{-4mm}
\label{fig1}
\end{figure}

%In today's technologically driven world, where multimedia content and virtual communication dominate, the ability to understand who is speaking is paramount to understanding conversations in context. 
% \south{In the aftereffect of the COVID-19 pandemic, the world is rapidly undergoing digitalisation \cite{pandey2020impact}}. 

% With a recent shift in the way we communicate from in-person to remote, the importance of 

In recent years, there has been a shift in the way we communicate, transitioning from in-person exchanges to audio-visual interactions online.
With this shift in paradigm, the task of identifying the active speaker has become crucial in order to enable effective communication and understanding of conversations in context.
% in paradigm  communication, the task of identifying the active speaker has become crucial in order to enhance communication efficiency and understand conversations in context.
% The objective of Active Speaker Detection (ASD) is to determine the active speaker in every video frame. 
In multi-modal conversations, active speaker detection (ASD) serves as a fundamental pre-processing module for speech-related tasks, including audio-visual speech recognition~\cite{afouras2018deep}, speech separation~\cite{owens2018audio,afouras2018conversation}, and speaker diarisation~\cite{chung2019said,chung2020spot}.

%ASD is frequently employed as a pre-processing step for speech-related tasks, including audio-visual speech recognition~\cite{afouras2018deep}, speech separation~\cite{owens2018audio,afouras2018conversation}, and speaker diarization~\cite{chung2019said,chung2020spot}.

ASD in real-world scenarios is a challenging task that requires effective integration of audio-visual information as well as leveraging their long-term relationships.
In response to these specific requirements, ASD model architectures are designed to create corresponding audio-visual features and comprehensively analyze these features over long periods to capture crucial temporal context.
The general ASD frameworks begin with modality-specific encoders extracting embedding and subsequent audio-visual fusion techniques enable the seamless interaction of heterogeneous modalities.
A prevalent fusion method involves the utilization of a cross-attention mechanism~\cite{talknet, loconet, jiang2023target}, facilitating the connection between visual streams and synchronized audio streams.
After the two features are combined, self-attention~\cite{talknet, loconet} or RNN-based architectures~\cite{lightASD, lwt} are employed to ensure consistent tracking of the active speaker throughout the utterance.
These aforementioned approaches take advantage of the complementary information of both modalities. 
However, learning quality representations for the multi-modal task of ASD has been less explored.
% it is worth noting that there has been limited effort directed toward enhancing the quality of learnt representations which is important for multi-modal applications. 

\begin{figure*}[!h]
\centering
\includegraphics[width=16cm]{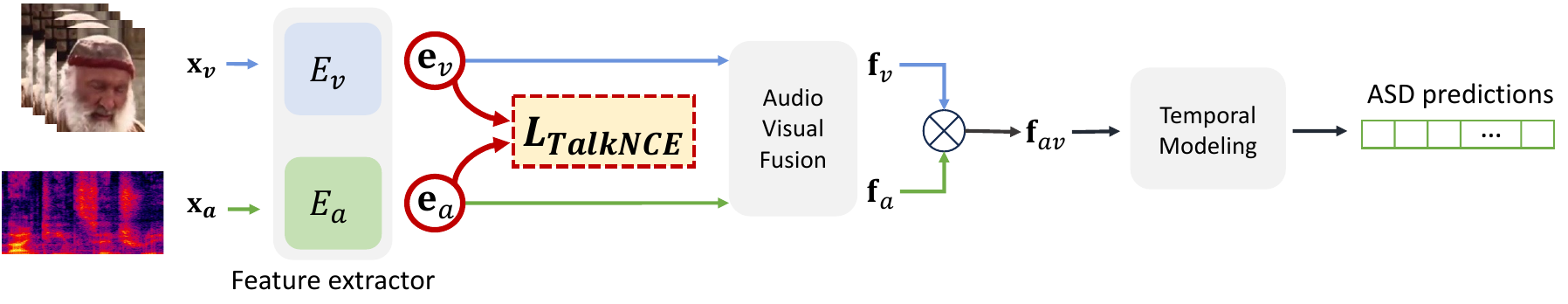}
\vspace{-2mm}
\caption{General framework for the active speaker detection task. Our loss is applied to the audio and visual embedding before the fusion. $\bigotimes$ denotes the concatenation of two features along the temporal dimension.}
\vspace{-1mm}
\label{fig2}
\end{figure*}

In this paper, we focus on learning strong phonetic representations for the ASD task.
To this end, we propose a novel supervised talk-aware contrastive learning strategy by devising a new loss function, named TalkNCE loss.
The loss acts on audio and visual features in a frame-wise manner rather than a chunk-wise manner~\cite{chung2017out,chung2020perfect} to learn transient phonetic representations required for audio-visual matching. 
Furthermore, the suggested loss is computed exclusively on the active speaking section extracted by using the ASD labels.
Supervision imposed by the TalkNCE loss enables the encoders to concentrate on the fine details of audio-visual correspondence, thus enhancing the effect of audio-visual fusion.
In contrast to the approaches that use pre-trained audio-visual embeddings~\cite{lwt, zhang2019multi}, our models are trained in an end-to-end manner by combining the proposed TalkNCE loss with the existing ASD classification losses.
Our contrastive learning strategy is model-agnostic, on many of which the addition of the TalkNCE loss brings performance improvements as shown in Fig.~\ref{fig1}.
In particular, combined with \cite{loconet}, our method outperforms the previous state-of-the-art ASD method on the AVA-ActiveSpeaker and ASW datasets.

Our contributions can be summarized as follows:
(1) We propose TalkNCE loss, a novel contrastive loss for ASD that enforces the model to exploit information from the alignment of audio and visual streams. 
(2) The proposed loss can be plugged into multiple existing ASD frameworks and the additional supervision imposed by the loss improves the performances without any additional data.
(3) Our method achieves state-of-the-art performances on AVA-ActiveSpeaker and ASW datasets.

\section{Method}
\label{sec:method}

\subsection{Preliminaries}
\label{ssec:method_prelim}

This section describes the general ASD framework widely used in literature~\cite{roth2020ava, talknet, lightASD, loconet, asc}.
As shown in Fig.~\ref{fig2}, ASD frameworks usually consists of feature extractors, audio-visual fusion, and temporal modeling. 
Given the input video frames, every face is detected, cropped, stacked, and transformed to grayscale images $\mathbf{x}_v$. The corresponding audio waveform is transformed into mel-spectrogram $\mathbf{x}_a$ by the short-time Fourier transform.
Then the audio encoder $E_a(\cdot)$ and visual encoder $E_v(\cdot)$ encode inputs into audio and visual embeddings respectively,
\begin{equation}
\mathbf{e}_a\!=\!E_a (\mathbf{x}_a), \quad \mathbf{e}_v\!=\!E_v (\mathbf{x}_v)
\end{equation}

To integrate the information between two modalities, the two features are fed into an audio-visual fusion module. 
Recent methods such as TalkNet~\cite{talknet} and LoCoNet~\cite{loconet} utilize cross-attention layers for multi-modal fusion. 
By cross-attention mechanism, audio embedding is effectively aligned with the visual information of the corresponding speaker.
Then the attention-weighted features $\mathbf{f}_a$ and $\mathbf{f}_v$ are concatenated to make the fused audio-visual features $\mathbf{f}_{av}$.

%%%% 3) Temporal modeling
Finally, several works~\cite{talknet, loconet, lightASD, EASEE, spell} employ a temporal modeling module by utilizing longer context to accurately predict the active speaker. 
% Especially, LoCoNet~\cite{loconet} utilizes a combination of self-attention for a long-term intra-speaker analysis and convolutional layers to take account of short-term inter-speaker contexts at the same time. 
Including self-attention modules in LoCoNet~\cite{loconet}, a Gated Recurrent Unit (GRU), a Long Short-Term Memory (LSTM), and a Bidirectional LSTM (BiLSTM) are frequently exploited for temporal modeling.

\begin{figure}[!t]
\vspace{-3mm}
  \centering
  \includegraphics[width=0.9\linewidth]{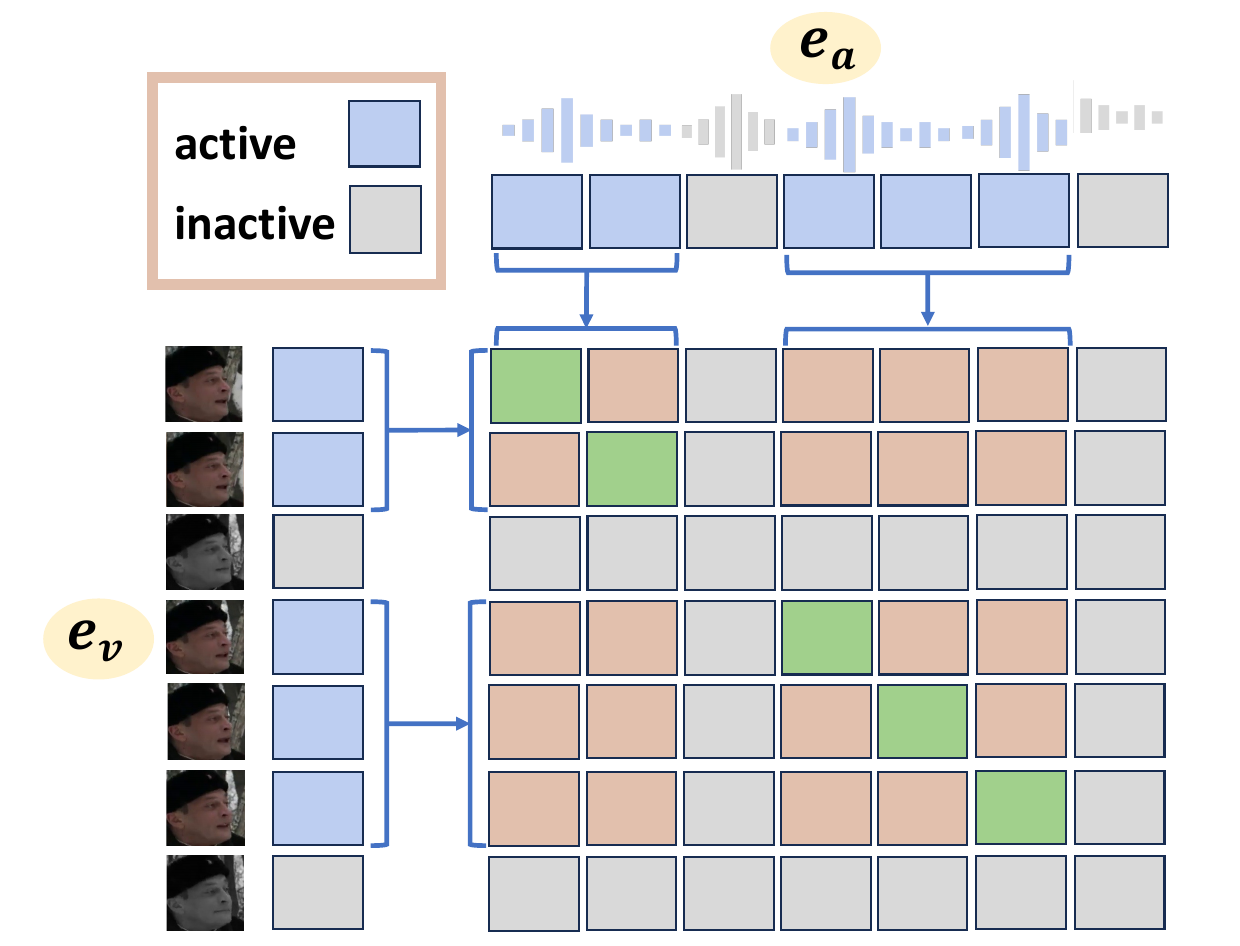}
% \caption{Our TalkNCE loss is applied only to active speaking section (marked as blue) in a frame-wise manner. Audio and visual embeddings of synchronized video frames are positive pairs (marked as green), while others are negative pairs (marked as orange).} 
\caption{Our TalkNCE loss is applied only to active speaking sections (blue) in a frame-wise manner. Audio and visual embeddings of synchronized video frames are positive pairs (green), while others are negative pairs (orange).} 
\vspace{-4mm}
\label{fig:res}
\end{figure}

\subsection{Contrastive Learning with TalkNCE Loss}
\label{ssec:method_loss}

The training objective typically used in the ASD task~\cite{roth2020ava, talknet, lightASD, loconet, asc} can be represented as: 
\begin{equation} 
\label{eq2}
 \mathcal{L}_{model} = \mathcal{L}_{av} + \lambda_a \cdot \mathcal{L}_a + \lambda_v \cdot \mathcal{L}_v, 
\end{equation}
where $\mathcal{L}_{av}$ denotes the cross-entropy loss between the ground-truth labels and the final frame-level predictions after temporal modeling. 
$\mathcal{L}_a$ and $\mathcal{L}_v$ are auxiliary cross-entropy losses that are computed with uni-modal features $\mathbf{f}_a$ and $\mathbf{f}_v$ to make the model utilize both modalities in a balanced manner.

In addition to the cross-entropy loss, we introduce TalkNCE loss, a talk-aware contrastive loss that provides supervision for encoders to learn audio-visual correspondence.
The proposed loss brings audio-visual embedding pairs closer if they come from the same video frame or pushes each other away otherwise. 
This encourages the encoders to detect salient information regarding audio-visual synchronization and to concentrate on phonetically fine details including frame-level correlation between audio and visual modalities.

% zhang2019multi, wuerkaixi2022rethinking: includes whole segments
% lwt, chung2017out: requires pre-training of encoders
Our new loss optimizes frame-level matching between paired audio and visual streams as denoted in Fig.~\ref{fig:res}. 
Given the audio and visual embeddings $\mathbf{e}_a$ and $\mathbf{e}_v$, the active speaking regions are extracted using the original ASD labels.
Let $T_{act}$ be the length of active speaking region and $\{\mathbf{e}_{a,i}\}_{i \in \{1,...,T_{act}\}}$, $\{\mathbf{e}_{v,i}\}_{i \in \{1,...,T_{act}\}}$ be the frame-level audio and visual embeddings for the active speaking region.
Then the TalkNCE loss $\mathcal{L}_{TalkNCE}$ is as follows:
\begin{equation}
\label{eq4}
    \mathcal{L}_{TalkNCE} = -\cfrac{1}{T_{act}} \sum_{i=1}^{T_{act}} \log \cfrac{\exp(s_{i,i} \slash \tau)}{\sum_{j=1}^{T_{act}} \mathbbm{1}_{[j \neq i]} \exp(s_{i,j} \slash \tau)},
\end{equation}
where $s_{i,j} = \norm{\mathbf{e}_{v,i}}^T \norm{\mathbf{e}_{a,i}}$ and $\tau$ represents a temperature constant fixed to 1. 
% Summing up the TalkNCE loss and the original ASD loss, t
With our proposed TalkNCE loss and the ASD loss, the final training objective used in this paper can be formulated as follows:
\begin{equation}
\label{eq5}
\begin{split}
 %  \mathcal{L} = &\, \mathcal{L}_{av} + \lambda_a \cdot \mathcal{L}_a + \lambda_v \cdot \mathcal{L}_v & +\lambda_{TalkNCE} \cdot \mathcal{L}_{TalkNCE},
   \mathcal{L} = \mathcal{L}_{model} +\lambda \cdot \mathcal{L}_{TalkNCE},
\end{split}
\end{equation}
where $\lambda$ is the weight value of TalkNCE loss.

\section{Experiments}
\label{sec:experiments}

% \begin{table}[!t]
% \centering
% \resizebox{1.0\linewidth}{!}{
% \renewcommand{\arraystretch}{1.2}
% \begin{tabular}{lc} 
% \Xhline{2\arrayrulewidth}
% \Xhline{0.1pt}  
% Method                      & mAP(\%)$\uparrow$        \\ 
% \hline
% ASC~\cite{asc}              & 87.1           \\
% MAAS~\cite{maas}               & 88.8           \\
% TalkNet~\cite{talknet}           & 92.3           \\
% ASDNet ~\cite{kopuklu2021design}       & 93.5           \\
% EASEE-50 ~\cite{EASEE}           & 94.1           \\
% \Xhline{2\arrayrulewidth}
% \Xhline{0.1pt}  
% \end{tabular}
% \begin{tabular}{lc}
% \Xhline{2\arrayrulewidth}
% \Xhline{0.1pt} 
% Method                      & mAP(\%)$\uparrow$        \\ 
% \hline
% SPELL ~\cite{spell}               & 94.2           \\
% TS-TalkNet ~\cite{jiang2023target}  & 93.9           \\
% Light-ASD~\cite{lightASD}           & 94.1           \\
% LoCoNet~\cite{loconet}      & 95.2           \\
% \hline
% \textbf{Ours}            & \textbf{95.5}  \\
% \Xhline{2\arrayrulewidth}
% \Xhline{0.1pt}  
% \end{tabular}
% }
% \vspace{-2mm}
% \caption{Performance comparison for previous methods and ours on the AVA-ActiveSpeaker validation set in terms of mAP. Performance of previous methods are from \cite{lightASD}.}
% \label{table1}
% \end{table}

\begin{table}[!t]

\centering
\resizebox{1.0\linewidth}{!}{
\renewcommand{\arraystretch}{1.2}
\begin{tabular}{lccc} 
% \Xhline{2\arrayrulewidth}
% \Xhline{0.1pt}  
\toprule
Method            & Multiple candidates?     & E2E?     & mAP(\%)        \\ 
% \hline
\midrule
ASC$^{\ast}$~\cite{asc}                    &   \cmark     &  \xmark  & 87.1           \\
MAAS$^{\ast}$~\cite{maas}                  &   \cmark     &  \xmark  & 88.8           \\
TalkNet$^{\ast}$~\cite{talknet}            &   \xmark     &  \cmark  & 92.3           \\
ASDNet$^{\ast}$~\cite{kopuklu2021design}  &   \cmark     &  \xmark  & 93.5           \\
EASEE-50$^{\ast}$~\cite{EASEE}            &  \cmark      &  \cmark  & 94.1           \\
SPELL$^{\ast}$~\cite{spell}               &  \cmark      &  \xmark  & 94.2           \\
Light-ASD$^{\ast}$~\cite{lightASD}         &  \xmark      &  \cmark  & 94.1           \\
TS-TalkNet~\cite{jiang2023target}         &  \xmark      &  \xmark  & 93.9           \\
LoCoNet~\cite{loconet}                     &  \cmark      &  \cmark  & 95.2           \\
\hline
\textbf{Ours}                              &  \cmark      &  \cmark  & \textbf{95.5}  \\
% \Xhline{2\arrayrulewidth}
% \Xhline{0.1pt}  
\bottomrule
\end{tabular}
}
\vspace{-2mm}
\caption{Comparison of ASD performance on the AVA-ActiveSpeaker validation set. $^{\ast}$Performance of previous methods are from \cite{lightASD}. E2E refers to end-to-end training.}
\vspace{-3mm}
\label{table1}
\end{table}

\subsection{Datasets}
\label{ssec:exp_datasets}
\newpara{AVA-ActiveSpeaker dataset}~\cite{roth2020ava} is a benchmark for evaluating active speaker detection performance extracted from Hollywood movies. It contains 262 videos that span 38.5 hours, of which 120, 33, and 109 videos are divided into training, validation, and test set respectively. As the videos are from movies, the dataset includes some dubbed videos.
%, and `active' labels are used even though the audio and video are not perfectly synchronized.

\newpara{Active Speakers in the Wild (ASW) dataset}~\cite{lwt} is a dataset derived from the existing audio-only speaker diariazation dataset VoxConverse~\cite{chung2020spot} by using a subset of the videos and exploiting the diarization annotations.
Unlike AVA-ActiveSpeaker, ASW excludes dubbed videos to ensure synchronization between lip movements and speech. Videos in the dataset total 30.9 hours, and the ratio of an active track duration is 56.7\%, 60.4\%, and 57.0\% for the training, validation, and test set respectively. Note that 4 videos cannot be downloaded since the videos are no longer online.

\subsection{Experimental Setup}
\label{ssec:exp_impl}

\newpara{Implementation details.}
Visual input $\mathbf{x}_v \in \mathbbm{R}^{n_s \times T \times H \times W}$ is pre-processed to size of $H=112$ and $W=112$ for $n_s$ speakers and $T$ video frames. The corresponding audio signal is transformed into mel-spectrogram $X_a \in \mathbb{R}^{4T*C_{mel}}$. The mel-spectrogram is extracted to have $4T$ time frames, by adjusting the hop length of short-time Fourier transform.
The frame-level audio embedding $e_a \in \mathbb{R}^{T \times C}$ and visual embedding $e_v \in \mathbb{R}^{T \times C}$ are extracted with a feature size of $C=128$.

We reproduce the baselines for performance comparison following the official reporting of each model~\cite{loconet,lightASD, talknet}. 
We train each model end-to-end for 25 epochs using the Adam optimizer~\cite{KingBa15}. 
Feature dimensions of both audio and visual embeddings are set to 128 for all three models. 
The value for $\lambda$ is set to 0.3 for LoCoNet to make the scale similar with $\mathcal{L}_{av}$. 
Similarly, $\lambda$ values for other models\cite{talknet, lightASD} are assigned considering the magnitudes of the existing loss function.

\newpara{Evaluation metrics.}
Following the common protocol suggested by previous works~\cite{loconet, lightASD, talknet, kopuklu2021design}, mean Average Precision (mAP) is used as an evaluation metric for the AVA-ACtiveSpeaker validation set. 
Evaluation metrics reported for the ASW validation and test sets are mAP, Area Under the ROC Curve (AUC), and Equal Error Rate (EER).
% As a multi-speaker model, outputs of LoCoNet are the predictions of not only target speakers but also context speakers.
% Therefore, the original loss is an average of cross entropy loss values computed for the prediction of every speaker.
% Following this, TalkNCE loss is also computed using audio and visual embeddings of active speaking frames for all speakers.
% For the other models such as TalkNet~\cite{talknet} which exploits only one speaker's information, we follow the training details of each model when our loss is applied. 

\begin{table}[!t]
\centering
% \resizebox{0.8\linewidth}{!}{
\resizebox{0.6\linewidth}{!}{
\renewcommand{\arraystretch}{1.2}
\begin{tabular}{lc} 
% \Xhline{2\arrayrulewidth}
% \Xhline{0.1pt}  
\toprule
Method           &   mAP(\%)            \\ 
% \hline
\midrule
TalkNet$^{\dagger}$               & 92.0              \\
TalkNet+$ \mathcal{L}_{TalkNCE}$     & 92.5        \\
LightASD$^{\dagger}$                 & 93.9                \\
LightASD+$\mathcal{L}_{TalkNCE}$  & 94.2   \\
LoCoNet$^{\dagger}$                  &95.2               \\
\textbf{LoCoNet+$\mathcal{L}_{TalkNCE}$} & \textbf{95.5}   \\
% \Xhline{2\arrayrulewidth}
% \Xhline{0.1pt}  
\bottomrule
\end{tabular}
}
\vspace{-2mm}
\caption{Performance comparison of other baseline models trained with TalkNCE Loss. $^{\dagger}$ Results are reproduced using the codes released by the original works.}
\label{table2}
\end{table}
\subsection{Performance on AVA-ActiveSpeaker Dataset}
\label{ssec:exp_ava}
The performance of the our method is compared with that of existing ASD methods on the AVA-ActiveSpeaker validation set, and we show our results in Table~\ref{table1}.
% Table~\ref{table1} shows the comparison of our method and previous results.
As indicated in Table~\ref{table1}, LoCoNet trained with our method attains 95.5\% mAP, achieving a state-of-the-art result.
Note that the model can be trained in an end-to-end manner using the combination of the proposed contrastive loss with the cross-entropy loss, without the need for a multiple-stage training strategy.
We also apply our contrastive learning strategy for TalkNet and Light-ASD to verify the general capability of our method. 
It can be seen in Table~\ref{table2} that our loss provides a consistent increase in performances compared to our reproduced models~\cite{talknet, lightASD}.

%For TalkNet, our loss enhances the best mAP for 0.44\%p in Table~\ref{table2}. 
%For the TalkNet model trained along with our loss, the performance on the earlier epochs were higher than the later epochs of exiting TalkNet. 
%For the LightASD model, the loss has little effect on the rate of performance convergence, but there is no doubt that the loss increases the performance of the reproduced Light-ASD.

%It is evident that our loss increases performances in our reproduced Light-ASD.

%Other than the baseline, we also adopt the loss for TalkNet and Light-ASD. For TalkNet, our loss enhances the best mAP for 0.44\%p in Table~\ref{table2}. \syun{For the TalkNet model trained along with our loss, the performance at earlier epochs are higher than the performance at later epochs of the vanilla TalkNet. For the LightASD model, the loss had little effect on the rate of performance convergence, but there is no doubt that the loss increases the performance of the reproduced Light-ASD.}

\subsection{Performance on ASW Dataset}
\label{ssec:exp_asw}

We also compare our method with other ASD methods on the ASW dataset.
In Table~\ref{table3}, our method achieves state-of-the-art results by a margin of 1.1\% on the ASW validation set and 0.8\% for the test set.
% Table~\ref{table3} is constructed based on \cite{jiang2023target}.
% In addition, for a fair comparison on the ASW dataset, we use an evaluation script provided by the author of \cite{jiang2023target}. 
% The highest performance is obtained by applying our talk-aware contrastive loss 
Applying our talk-aware contrastive loss $\mathcal{L}_{TalkNCE}$ to LoCoNet shows the highest performance. 

In contrast to the AVA-ActiveSpeaker dataset, the ASW dataset does not include any dubbed video. This makes our loss more effective for the ASW dataset since our loss is devised to optimize the performance through learning audio-visual phonetic representations for synchronization.
% There is no dubbed video in ASW dataset different from AVA dataset which involves dubbed videos. Since TalkNCE loss is applied to audio-visual synchronized regions, our loss can press enormous effects on ASW performances.

% val (mAP = 98.81, AUC = 99.34, EER = 3.17 @ epoch22)
% test(mAP = 99.30, AUC = 99.52, EER = 2.99 @ epoch19)

\begin{table}[!t]
\centering
    \resizebox{0.96\linewidth}{!}{%
    \begin{tabular}{l|c|ccc} 
    % \Xhline{2\arrayrulewidth}
    % \Xhline{0.1pt}  
    \toprule
    &Method                      & mAP(\%)$\uparrow$  & AUC(\%)$\uparrow$ & EER(\%)$\downarrow$      \\ 
    % \hline
    \midrule
    \multirow{3}{*}{Val} &TalkNet~\cite{talknet} & 96.4 & 98.2& 6.0\\
    &TS-TalkNet~\cite{jiang2023target} &97.7& 98.7& 5.1\\
    &\textbf{Ours} & \textbf{98.8} & \textbf{99.3} & \textbf{3.2} \\
    % \hline
    \midrule
    \multirow{5}{*}{Test}&TalkNet~\cite{talknet} & 97.7 & 98.6 & 5.1  \\
    &TS-TalkNet~\cite{jiang2023target} &98.5 &99.0 &4.3 \\
    &ASW-BGRUs~\cite{lwt} & 96.6 & 97.2 & 6.2\\
    &LoCoNet~\cite{loconet} & 93.4& 95.1& 9.8\\
    &\textbf{Ours} & \textbf{99.3} & \textbf{99.5} & \textbf{3.0} \\
    % \Xhline{2\arrayrulewidth}
    % \Xhline{0.1pt}  
    \bottomrule
    \end{tabular}%
    }
\vspace{-2mm}
\caption{Comparison of ASD performance on the ASW dataset. Performance of previous methods are from \cite{jiang2023target}. $\uparrow$ denotes higher is better, and $\downarrow$ denotes lower is better.}
\label{table3}
\vspace{-2mm}
\end{table}

% val (mAP = 98.81, AUC = 99.34, EER = 3.17 @ epoch22)
% test(mAP = 99.30, AUC = 99.52, EER = 2.99 @ epoch19)

\subsection{Ablation Studies}
\label{ssec:exp_ablation}

\newpara{Choice of positive and negative samples.}
By using the ASD labels for a certain speaker, a sequence of audio and visual embeddings can be divided into `active' and `inactive' regions, as shown in Fig.~\ref{fig:res}. 
Our proposed method only attends to the `active' regions for calculating the contrastive loss $\mathcal{L}_{TalkNCE}$, and Table~\ref{table_abl_samples} shows the results obtained by different sampling strategies. 
% In Table~\ref{table_abl_samples}, `active' means that $\mathcal{L}_{TalkNCE}$ is calculated for only active regions for the embedding, and `all' means that $\mathcal{L}_{TalkNCE}$ is calculated by utilizing all regions regardless of whether active or not. 
The experimental result demonstrates that the model can learn richer meaningful representations from the active part of the inputs, rather than from the inactive part.
In particular, utilizing inactive frames of audio embedding degrades performance, as the audio in inactive regions can be irrelevant to the speaker's lip motions. 
Similarly, the visual input from inactive regions contains non-speaking faces, which can have an adverse effect on discriminative learning. 
% Visual information from inactive regions, such as silent faces, can be classified as a negative pair with audio from active regions. 
% Non-speaking lip movements in inactive video regions can complicate the accurate synchronization of audio and video, resulting in lower performance compared to the combination of active regions.

%
% Utilizing the full segment increases the total number of training samples for our loss, but silent faces and irrelevant audio are paired as positive, which is not desirable to focus on the correspondence between the two modalities. Therefore, this approach also degrades performance, and we show that our loss can optimize the model when the accurate selection of active pairs is provided.

\newpara{Location of $\mathcal{L}_{TalkNCE}$.} % Seeking the location of $L$ for propagation
Our proposed loss is particularly effective for refining the audio and visual embeddings before audio-visual fusion as indicated in Table~\ref{table_abl_lambda}. 
When it is applied after the audio-visual fusion stage, it has a negative effect on performance.
Applying contrastive learning to the attention-weighted features $\mathbf{f}_a$ and $\mathbf{f}_v$ hinders the effect of audio-visual fusion.
%seems to have a negative influence on performances. \jyj{(see comment.)} 
As our TalkNCE loss can refine uni-modal features by giving information from other modalities using frame-level contrastive loss, our loss could negatively impact $\mathbf{f}_a$ and $\mathbf{f}_v$ operating on multi-modal guidance, which have already interacted with each other during the fusion stage.

%following concatenation operation than cross-attention, even harming the performance of the baseline. 

%so the performance declines. 
%between the relative audio-visual pair 

\newpara{Weight value of $\mathcal{L}_{TalkNCE}$.}
We also show the performance of LoCoNet trained using our loss with varying $\lambda$ values for $\mathcal{L}_{TalkNCE}$.
As shown in Table~\ref{table_abl_lambda}, the $\lambda$value of 0.3 gives the best performance.
% while its value does not have a huge impact on performance.
Our loss and the ASD classification loss $\mathcal{L}_{av}$ can be trained in balance at this value.

% \begin{table}[!t]
% \centering
% \resizebox{0.9\linewidth}{!}{
% \begin{tabular}{cc|cc|c}
%  \Xhline{2\arrayrulewidth}
%  \Xhline{0.1pt} 
% % \multicolumn{2}{c|}{$\mathbf{e}_v$} & \multicolumn{2}{c|}{$\mathbf{e}_a$} & \multirow{2}{*}{mAP(\%)$\uparrow$} \\ \cline{1-4}
% \multicolumn{2}{c|}{Visual emb} & \multicolumn{2}{c|}{Audio emb} & \multirow{2}{*}{mAP(\%)} \\ \cline{1-4}
% Active     & Inactive     & Active     & Inactive     &                          \\ \hline
% \checkmark         & \checkmark           & \checkmark         & \checkmark           & 94.9                     \\
% \checkmark         &              & \checkmark         & \checkmark           & 94.3                     \\
% \checkmark         & \checkmark           & \checkmark         &              & 95.1                     \\
% \checkmark         &              & \checkmark         &              & \textbf{95.5 }            \\       
% \Xhline{2\arrayrulewidth}
% \Xhline{0.1pt} 
% \end{tabular}
% }
% \vspace{-2mm}
% \caption{Performance comparison of 4 combinations of visual and audio embeddings used for calculating the TalkNCE loss. 
% }
% \label{table_abl_samples}
% \end{table}
%------------------------------------------------------------------------
\begin{table}[t!]
\centering
\resizebox{0.85\linewidth}{!}{
\label{table4}
\begin{tabular}{llc}
% \Xhline{2\arrayrulewidth}
% \Xhline{0.1pt}  
\toprule
Visual embedding & Audio embedding  & mAP(\%)      \\ 
\midrule
Act & Act        & \textbf{95.5}  \\
Act & Act + Inact    & 94.3  \\
Act + Inact & Act      & 95.1  \\
Act + Inact & Act + Inact   & 94.9  \\
% \Xhline{2\arrayrulewidth}
% \Xhline{0.1pt}  
\bottomrule
\end{tabular}
}

\caption{Comparisons between combinations of visual and audio embeddings used for calculating our TalkNCE loss. `Act' refers to the active region, and `Inact' refers to inactive regions for a certain speaker.}
\label{table_abl_samples}
\end{table}

\begin{table}[!t]
% \centering
\resizebox{.9\linewidth}{!}{
\begin{minipage}{0.5\linewidth}
        \begin{tabular}{lc} 
% \Xhline{2\arrayrulewidth}
% \Xhline{0.1pt}  
\toprule
Location of $\mathcal{L}_{TalkNCE}$   & mAP(\%)       \\ 
% \hline
\midrule
None (Baseline) & 95.2 \\
After CA           & 93.4            \\
\textbf{Before CA}    & \textbf{95.5}  \\
% \Xhline{2\arrayrulewidth}
% \Xhline{0.1pt}  
\bottomrule
\end{tabular}
\end{minipage}
\hspace{1.35cm}
\begin{minipage}{0.25\linewidth}
    % \centering
    % \hspace*{\fill}
        \begin{tabular}{cc} 
        % \Xhline{2\arrayrulewidth}
        % \Xhline{0.1pt}  
        \toprule
     $ \lambda$                & mAP(\%)       \\ 
    % \hline
    \midrule
    0.15         & 95.2            \\
    \textbf{0.3}           & \textbf{95.5}        \\
    0.6    & 95.2  \\
    % \Xhline{2\arrayrulewidth}
    % \Xhline{0.1pt}  
    \bottomrule
    \end{tabular}
\end{minipage}
}

\caption{Ablation on the position and the $\lambda$ value of the TalkNCE loss, using LoCoNet~\cite{loconet} as a baseline. CA denotes the cross-attention module for audio-visual fusion.}
% Ablation results of LoCoNet~\cite{loconet}. Table (a) shows results regarding different $\lambda$ values. Table (b) shows results according to the position of TalkNCE loss. CA denotes the cross-attention module inside LoCoNet for audio-visual fusion.}

\label{table_abl_lambda}
\end{table}

% Active frame region in which audio and visual features are synchronized as `active' in Fig~\ref{fig:res}. The last row in Table~\ref{table_abl_lambda} shows the highest score in AVA-ActiveSpeaker validation dataset. The other rows are results of the ablation studies on various combinations of active and inactive regions.
% In active regions with an audio modality, visual embeddings whether active or not can be utilized as negative pairs, as speaker identities in both regions are same. 
% As audio embeddings in non-active region contain irrelevant sounds for the speaker, such as silence or another speaker's talk, we may treat the audio feature in the non-active frame as an additional negative sample.
% Moreover, we can utilize all samples to increase the number of samples. However, above approaches degrade performances.
% We guess that it is not directly related to the scores to use various samples because our loss needs delicate selection of positive and negative pairs to deal with synchronization between multi-modalities. 

\section{Conclusion}
\label{sec:typestyle}

In this paper, we propose TalkNCE, a talk-aware contrastive loss for active speaker detection. The objective function utilizes active speaker labels to select audio and visual embeddings from which natural co-occurrences are learnt. Our loss is applicable to a range of existing ASD systems, for which we demonstrate a consistent improvement in performance. In particular, we achieve a new state-of-the-art score of 95.5\% on the AVA-ActiveSpeaker validation set by combining the proposed method with the LoCoNet model.

\clearpage
\vfill\pagebreak

% -------------------------------------------------------------------------
\bibliographystyle{IEEEbib}
\bibliography{shortstrings,refs}

\end{document}